# Incremental Pruning: A Simple, Fast, Exact Method for Partially Observable Markov Decision Processes

**Anthony Cassandra**
Computer Science Dept.
Brown University
Providence, RI 02912
arc@cs.brown.edu

**Michael L. Littman**
Dept. of Computer Science
Duke University
Durham, NC 27708-0129
mlittman@cs.duke.edu

**Nevin L. Zhang**
Computer Science Dept.
The Hong Kong U. of Sci. & Tech.
Clear Water Bay, Kowloon, HK
lzhang@cs.ust.hk

## Abstract

Most exact algorithms for general partially observable Markov decision processes (POMDPs) use a form of dynamic programming in which a piecewise-linear and convex representation of one value function is transformed into another. We examine variations of the "incremental pruning" method for solving this problem and compare them to earlier algorithms from theoretical and empirical perspectives. We find that incremental pruning is presently the most efficient exact method for solving POMDPs.

## 1 INTRODUCTION

Partially observable Markov decision processes (POMDPs) model decision theoretic planning problems in which an agent must make a sequence of decisions to maximize its utility given uncertainty in the effects of its actions and its current state (Cassandra, Kaelbling, & Littman 1994; White 1991). At any moment in time, the agent is in one of a finite set of possible states $S$ and must choose one of a finite set of possible actions $\mathcal{A}$. After taking action $a \in \mathcal{A}$ from state $s \in S$, the agent receives immediate reward $r^a(s) \in \Re$ and the agent's state becomes some state $s'$ with the probability given by the transition function $\Pr(s'|s,a) \in [0,1]$. The agent is not aware of its current state, and instead only knows its information state $x$, which is a probability distribution over possible states ($x(s)$ is the probability that the agent is in state $s$). After each transition, the agent makes an observation $z$ of its current state from a finite set of possible observations $\mathcal{Z}$. The function $\Pr(z|s',a) \in [0,1]$ gives the probability that observation $z$ will be made after the agent takes action $a$ and makes a transition to state $s'$. This results in a new information state $x_z^a$ defined by

$$x_z^a(s') = \frac{\Pr(z|s',a)\sum_{s \in S}\Pr(s'|s,a)x(s)}{\Pr(z|x,a)}, \quad (1)$$

where

$$\Pr(z|x,a) = \sum_{s' \in S}\Pr(z|s',a)\sum_{s \in S}\Pr(s'|s,a)x(s).$$

Solving a POMDP means finding a policy $\pi$ that maps each information state into an action so that the expected sum of discounted rewards is maximized ($0 \leq \gamma \leq 1$ is the discount rate, which controls how much future rewards count compared to near-term rewards). There are many ways to approach this problem based on checking which information states can be reached (Washington 1996; Hansen 1994), searching for good controllers (Platzman 1981), and using dynamic programming (Smallwood & Sondik 1973; Cheng 1988; Monahan 1982; Littman, Cassandra, & Kaelbling 1996).

Most exact algorithms for general POMDPs use a form of dynamic programming in which a piecewise-linear and convex representation of one value function is transformed into another. This includes algorithms that solve POMDPs via value iteration (Sawaki & Ichikawa 1978; Cassandra, Kaelbling, & Littman 1994), policy iteration (Sondik 1978), accelerated value iteration (White & Scherer 1989), structured representations (Boutilier & Poole 1996), and approximation (Zhang & Liu 1996). Because dynamic-programming updates are critical to such a wide array of POMDP algorithms, identifying fast algorithms is crucial.

Several algorithms for dynamic-programming updates have been proposed, such as one pass (Sondik 1971), exhaustive (Monahan 1982), linear support (Cheng 1988), and witness (Littman, Cassandra, & Kaelbling 1996). Cheng (1988) gave experimental evidence that the linear support algorithm is more efficient than the one-pass algorithm. Littman, Cassandra and Kaelbling (1996) compared the exhaustive algorithm, the linear support algorithm, and the witness algorithm and found that, except for tiny problems with approximately 2 observations or 2 states, which all three algorithms could solve quickly, witness was the fastest and had a number of superior theoretical properties.

Recently, Zhang and Liu (1996) proposed a new method for dynamic-programming updates in POMDPs called incremental pruning. In this paper, we analyze the basic algorithm and a novel variation and compare them to the witness algorithm. We find that the incremental-pruning-based algorithms allow us to solve problems that could not be solved within reasonable time limits using the witness algorithm.

## 2 DP UPDATES

The fundamental idea of the dynamic-programming (DP) update is to define a new value function $V'$ in terms of a given value function $V$. Value functions are mappings from information states to expected discounted total reward. In value-iteration algorithms, $V'$ incorporates one additional step of reward compared to $V$ and in infinite-horizon algorithms, $V'$ represents an improved approximation that is closer to the optimal value function.

The function $V'$ maps information states to values and is defined by

$$V'(x) = \max_{a \in \mathcal{A}} \left( \sum_{s \in \mathcal{S}} r^a(s) x(s) + \gamma \sum_{z \in \mathcal{Z}} \Pr(z|x, a) V(x_z^a) \right). \tag{2}$$

In words, Equation 2 says that the value for an information state $x$ is the value of the best action that can be taken from $x$ of the expected immediate reward for that action plus the expected discounted value of the resulting information state ($x_z^a$, as defined in Equation 1).

We can break up the value function $V'$ defined in Equation 2 into simpler combinations of other value functions:

$$V'(x) = \max_{a \in \mathcal{A}} V^a(x) \tag{3}$$

$$V^a(x) = \sum_z V_z^a(x) \tag{4}$$

$$V_z^a(x) = \frac{\sum_s r^a(s) x(s)}{|\mathcal{Z}|} + \gamma \Pr(z|x, a) V(x_z^a). \tag{5}$$

These definitions are somewhat novel and form an important step in the derivation of the incremental pruning method, described in Section 4. Each $V^a$ and $V_z^a$ function is a value function mapping information states to value and is defined in terms of relatively simple transformations of other value functions.

The transformations preserve piecewise linearity and convexity (Smallwood & Sondik 1973; Littman, Cassandra, & Kaelbling 1996). This means that if the function $V$ can be expressed as $V(x) = \max_{\alpha \in S} x \cdot \alpha$ for some finite set of $|\mathcal{S}|$-vectors $S$ (which it can in most applications), then we can express $V_z^a(x) = \max_{\alpha \in S_z^a} x \cdot \alpha$, $V^a(x) = \max_{\alpha \in S^a} x \cdot \alpha$, and $V'(x) = \max_{\alpha \in S'} x \cdot \alpha$ for some finite sets of $|\mathcal{S}|$-vectors $S_z^a$, $S^a$, and $S'$ (for all $a \in \mathcal{A}$, and $z \in \mathcal{Z}$). These sets have a unique representation of minimum size (Littman, Cassandra, & Kaelbling 1996), and we assume that the symbols $S_z^a$, $S^a$, and $S'$ refer to the minimum-size sets.

Here is a brief description of the set and vector notation we will be using. Vector comparisons are componentwise: $\alpha_1 \geq \alpha_2$ if and only if for all $s \in \mathcal{S}$, $\alpha_1(s) \geq \alpha_2(s)$. Vector sums are also componentwise. Vector dot products are defined by $\alpha \cdot \beta = \sum_s \alpha(s)\beta(s)$. In vector comparisons and dot products, 0 is a vector of all zeros and 1 a vector of all ones. For all $s \in \mathcal{S}$, the vector $e_s$ is all zeros except $e_s(s) = 1$. The *cross sum* of two sets of vectors is $A \oplus B = \{\alpha + \beta | \alpha \in A, \beta \in B\}$; this extends to collections of vector sets as well. Set subtraction is defined by $A \backslash B = \{\alpha \in A | \alpha \notin B\}$.

Using this notation, we can characterize the "$S$" sets described earlier as

$$S' = \text{purge}\left(\bigcup_{a \in \mathcal{A}} S^a\right) \tag{6}$$

$$S^a = \text{purge}\left(\bigoplus_{z \in \mathcal{Z}} S_z^a\right) \tag{7}$$

$$S_z^a = \text{purge}(\{\tau(\alpha, a, z) | \alpha \in S\}), \tag{8}$$

where $\tau(\alpha, a, z)$ is the $|\mathcal{S}|$-vector given by

$$\tau(\alpha, a, z)(s) = (1/|\mathcal{Z}|) r^a(s) + \gamma \sum_{s'} \alpha(s') \Pr(z|s', a) \Pr(s'|s, a),$$

and purge($\cdot$) takes a set of vectors and reduces it to its unique minimum form. Equations 6 and 7 are easily justified by Equations 3 and 4 and basic properties of piecewise-linear convex functions. Equation 8 comes from substituting Equation 1 into Equation 5, simplifying, and using basic properties of piecewise-linear convex functions.

The focus of this paper is on efficient implementations for computing $S^a$ (Equation 7). Equations 6 and 8 can be implemented efficiently using an efficient implementation of the purge function, described in the next section.

## 3  PURGING SETS OF VECTORS

Given a set of $|\mathcal{S}|$-vectors $A$ and a vector $\alpha$, define

$$R(\alpha, A) = \{x | x \geq 0, x \cdot 1 = 1, x \cdot \alpha > x \cdot \alpha', \alpha' \in A \setminus \{\alpha\}\}; \quad (9)$$

it is the set of information states for which vector $\alpha$ is the clear "winner" (has the largest dot product) compared to all the other vectors in $A$. The set $R(\alpha, A)$ is called the *witness region* of vector $\alpha$, because for any information state $x$ in this set $\max_{\alpha' \in A \setminus \{\alpha\}} x \cdot \alpha' \neq \max_{\alpha' \in A \cup \{\alpha\}} x \cdot \alpha'$; in a sense, $x$ can testify that $\alpha$ is needed to represent the piecewise-linear convex function given by $A \cup \{\alpha\}$.

Using the definition of $R$, we can define

$$\text{purge}(A) = \{\alpha | \alpha \in A, R(\alpha, A) \neq \emptyset\};$$

it is the set of vectors in $A$ that have non-empty witness regions and is precisely the minimum-size set for representing the piecewise-linear convex function given by $A$ (Littman, Cassandra, & Kaelbling 1996)[1].

Figure 1 gives an implementation of purge($F$)—given a set of vectors $F$, FILTER($F$) returns the vectors in $F$ that have non-empty witness regions, thereby "purging" or "filtering" or "pruning" out the unnecessary vectors. The algorithm is due to Lark (White 1991); Littman, Cassandra, & Kaelbling (1996) analyze the algorithm and describe the way that the argmax operators need to be implemented for the analysis to hold (ties must be broken lexicographically). The DOMINATE($\alpha, A$) procedure called in line 8 returns an information state $x$ for which $\alpha$ gives a larger dot product than any vector in $A$ (or $\bot$ if no such $x$ exists)—that is, it returns an information state in the region $R(\alpha, A)$. It is implemented by solving a simple linear program, illustrated in Figure 2.

The FILTER algorithm plays a crucial role in the incremental pruning method, so it deserves some additional explanation. The set $W$, initially empty, is filled with vectors $\omega$ that have non-empty witness regions $R(\omega, F)$; they are the "winners." Lines 3–5 find those winning vectors at the $e_s$ information states.

The "while" loop starting on line 6 goes through the vectors $\phi \in F$ one by one. For each, DOMINATE is used to see if there is an $x \in R(\phi, W)$. If there is not, we know $R(\phi, F)$ is empty, since $x \in R(\phi, F)$ implies $x \in R(\phi, W)$ since $W \subseteq F$. If DOMINATE finds an $x \in R(\phi, W)$, we add the winning vector (not necessarily $\phi$) at $x$ to $W$ and continue. Each iteration removes a vector from $F$, and when it is empty, every vector from $F$ will have been classified as either a winner or not a winner.

---

[1] This assumes that $A$ is a true set in that it contains no duplicate vectors.

FILTER($F$)
1  $W \leftarrow \emptyset$
2  for each $s$ in $\mathcal{S}$
3  do $\omega \leftarrow \text{argmax}_{\phi \in F} e_s \cdot \phi$
4     $W \leftarrow W \cup \{\omega\}$
5     $F \leftarrow F \setminus \{\omega\}$
6  while $F \neq \emptyset$
7  do $\phi \in F$
8     $x \leftarrow \text{DOMINATE}(\phi, W)$
9     if $x = \bot$
10       then $F \leftarrow F \setminus \{\phi\}$
11       else $\omega \leftarrow \text{argmax}_{\phi \in F} x \cdot \phi$
12            $W \leftarrow W \cup \{\omega\}$
13            $F \leftarrow F \setminus \{\omega\}$
14  return $W$

Figure 1: Lark's algorithm for purging a set of vectors.

DOMINATE($\alpha, A$)
1  $L \leftarrow \text{LP}(\text{variables:} x(s), \delta; \text{objective: max } \delta)$
2  for each $\alpha'$ in $A \setminus \{\alpha\}$
3  do ADDCONSTRAINT($L, x \cdot \alpha \geq \delta + x \cdot \alpha'$)
4  ADDCONSTRAINT($L, x \cdot 1 = 1$)
5  ADDCONSTRAINT($L, x \geq 0$)
6  if INFEASIBLE($L$)
7     then return ($\bot$)
8     else $(x, \delta) \leftarrow \text{SOLVELP}(L)$
9          if $\delta > 0$
10            then return ($x$)
11            else return $\bot$

Figure 2: Linear-programming approach to finding an information state in a vector's witness region.

### 3.1  USING PURGE IN DP

Given the FILTER procedure, it is trivial to compute the $S_z^a$ sets from $S$ and to compute $S'$ from the $S^a$ sets (Equations 8 and 6).

A straightforward computation of the $S^a$ sets from the $S_z^a$ sets (Equation 7) is also easy, and amounts to an exhaustive enumeration of all possible combinations of vectors followed by filtering the resulting sets. This algorithm is not efficient because the number of combinations of vectors grows exponentially in $|\mathcal{Z}|$. This can be a large number of vectors even when the $S^a$ sets are relatively small. This approach to computing the $S^a$ sets from the $S_z^a$ sets was essentially proposed by Monahan (1982) (under the name of "Sondik's one-pass algorithm").

## 3.2 COMPLEXITY ANALYSIS

We seek to express the running time of algorithms in terms of the number of linear programs they solve and the size of these linear programs. We choose this metric because all of the algorithms in this paper use linear programming as a fundamental subroutine (in the form of calls to DOMINATE($\alpha, A$)) and the solution of these linear programs is by far the most time-consuming part of the algorithms. In addition, traditional "operation count" analyses are cumbersome and unenlightening because of the difficulty of precisely characterizing the number of primitive operations required to solve each linear program.

We will express the running time of $W \leftarrow$ FILTER($F$) in terms of the size of the sets $F$ and $W$, the number of states $|S|$, and $m$, the number of vectors in $W$ that are found by checking the $e_s$ information states.

As is evident in Figure 1, each iteration of the "while" loop on line 6 removes one vector from $F$, and $m$ vectors are removed before the loop. This means the while loop is executed precisely $|F|-m$ times. Each iteration of the "while" loop makes a single call to DOMINATE, so there are $|F|-m$ linear programs solved in all cases. Each of these linear programs has one variable for each state in $\mathcal{S}$ and one for $\delta$. The total number of constraints in any one of these linear programs will be between $m+1$ and $|W|+1$. In the best case, the total number of constraints will be $|F|(m+1) - m|W| + |W|(|W|-1)/2 - m(m+1)/2$ and the worst case will have an additional $(|F|-|W|)(|W|-m)$ constraints.

When checking the $e_s$ information states, at least one vector in $W$ will be found. Further, when $|W| > 1$ we are guaranteed to find at least two of the vectors in $W$. For the remainder of this paper, we assume that $|W| > 1$, since the case of $|W| = 1$ is trivial.

The witness algorithm has been analyzed previously (Littman, Cassandra, & Kaelbling 1996), and we list the basic results here for easy comparison. The total number of linear programs solved by witness is $(\sum_z |S_z^a| - |\mathcal{Z}|)|S^a| + |S^a| - 1$; asymptotically, this is $\Theta(|S^a| \sum_z |S_z^a|)$. Note that this is not a worst-case analysis; this many linear programs will always be required. The number of constraints in each linear program is bounded by $|S^a| + 1$. The total number of constraints over all the linear programs is $\Theta(|S^a|^2 \sum_z |S_z^a|)$ asymptotically[2].

---

[2]In the best case there are $1/2(\sum_z |S_z^a| - |\mathcal{Z}| + 1)(|S^a| + 1)(|S^a| + 2) - \sum_z |S_z^a| + |\mathcal{Z}| - 3$ constraints and in the worst case there are $|S^a|(|S^a| + 1)(\sum_z |S_z^a| - |\mathcal{Z}| - 1/2)$ constraints.

INCPRUNE($S_{z_1}^a, \ldots, S_{z_k}^a$)
1  $W \leftarrow$ FILTER($S_{z_1}^a \oplus S_{z_2}^a$)
2  **for** $i \leftarrow 3$ **to** $k$
3  **do** $W \leftarrow$ FILTER($W \oplus S_{z_i}^a$)
4  **return** $W$

Figure 3: The incremental pruning method.

## 4 INCREMENTAL PRUNING

This section describes the incremental pruning method (Zhang & Liu 1996), which computes $S^a$ efficiently from the $S_z^a$ sets.

Recall the definition for $S^a$ in Equation 7:

$$S^a = \mathrm{purge}\left(\bigoplus_{z \in \mathcal{Z}} S_z^a\right) = \mathrm{purge}(S_{z_1}^a \oplus S_{z_2}^a \oplus \ldots \oplus S_{z_k}^a);$$

here, $k = |\mathcal{Z}|$. Note that

$$\mathrm{purge}(A \oplus B \oplus C) = \mathrm{purge}(\mathrm{purge}(A \oplus B) \oplus C),$$

so Equation 7 can be rewritten as

$$S^a = \mathrm{purge}(\ldots \mathrm{purge}(\mathrm{purge}(S_{z_1}^a \oplus S_{z_2}^a) \oplus S_{z_3}^a) \ldots \oplus S_{s_k}^a). \quad (10)$$

The expression for $S^a$ in Equation 10 leads to a very natural solution method, called incremental pruning, illustrated in Figure 3. In addition to being conceptually simpler than the witness algorithm, we will show that it can be implemented to exhibit superior performance and asymptotic complexity.

The critical fact required to analyze incremental pruning is that if $A = \mathrm{purge}(A)$ and $B = \mathrm{purge}(B)$ (neither contain extra vectors) and $W = \mathrm{purge}(A \oplus B)$, then

$$|W| \geq \max(|A|, |B|). \quad (11)$$

Equation 11 follows from the observation that for every $\omega \in W$, every $R(\omega, W)$ region is contained within $R(\alpha, A)$ and $R(\beta, B)$ for some $\alpha \in A$ and $\beta \in B$. This means that the size of the $W$ set in INCPRUNE is monotonically non-decreasing; it never grows explosively compared to its final size.

Figure 3 illustrates a family of algorithms that we collectively call the *incremental pruning method*; specific incremental pruning algorithms differ in their implementations of the FILTER procedure. The most basic incremental pruning algorithm is given by implementing FILTER by Lark's algorithm (Figure 1); we call the resulting algorithm IP. In Section 5, we describe several other variations.

The complexity of IP is $\Theta(|S^a|\sum_z |S_z^a|)$ linear programs and $O(|S^a|^2 \sum_z |S_z^a|)$ constraints[3]. In the worst case, these bounds are identical to those of the witness algorithm (Section 3.2). However, there are POMDPs for which the expression for the total number of constraints is arbitrarily loose; the best-case total number of constraints for IP is asymptotically better than for witness.

## 5   GENERALIZING IP

All the calls to FILTER in INCPRUNE (Figure 3) are of the form FILTER$(A \oplus B)$. This section modifies the implementation of FILTER to take advantage of the fact that the set of vectors being processed has a great deal of regularity. The modification yields a family of FILTER algorithms, some of which render incremental pruning more efficient than when the standard version appearing in Figure 1 is used.

The change is to replace line 8 in Figure 1 with

$$x \leftarrow \text{DOMINATE}(\phi, D \setminus \{\phi\}). \qquad (12)$$

Any set $D$ of vectors satisfying the properties below can be used and still give a correct algorithm (recall that we are filtering the set of vectors $A \oplus B$ and $W$ is the set of winning vectors found so far):

1. $D \subseteq (A \oplus B)$.

2. Let $(\alpha + \beta) = \phi$ for $\alpha \in A$ and $\beta \in B$. For every $\alpha_1 \in A$ and $\beta_1 \in B$, if $(\alpha_1 + \beta_1) \in W$, then either $(\alpha_1 + \beta_1) \in D$, or $(\alpha_1 + \beta) \in D$, or $(\alpha + \beta_1) \in D$.

There are a number of choices for $D$ that satisfy the above properties. For example,

$$D = A \oplus B, \qquad (13)$$
$$D = (\{\alpha\} \oplus B) \cup (A \oplus \{\beta\}), \qquad (14)$$
$$D = W, \qquad (15)$$
$$D = (\{\alpha\} \oplus B) \cup \{\alpha_1 + \beta | (\alpha_1 + \beta) \in W\}, (16)$$
$$D = (A \oplus \{\beta\}) \cup \{\beta_1 + \alpha | (\alpha + \beta_1) \in W\}, (17)$$

The following lemma shows that any such choice of $D$ allows us to use the domination check in Equation 12 to either remove $\phi$ from consideration, or to find a vector in purge$(A \oplus B)$ that has not yet been added to $W$ (note that $\phi \notin W$).

**Lemma 1** *If $R(\phi, D \setminus \{\phi\}) = \emptyset$, then $R(\phi, A \oplus B) = \emptyset$. If $x \in R(\phi, D \setminus \{\phi\})$, then $x \in R(\omega, W)$ for some $\omega \in (A \oplus B) \setminus W$.*

---

[3] Simple upper bounds on the IP algorithm are $|S^a|\sum_z |S_z^a|$ linear programs and $|S^a|(|S^a|+1)\sum_z |S_z^a| - 3|\mathcal{Z}|$ total constraints. Note that tighter, though more complicated, upper bounds are possible.

**Proof**: First, if $R(\phi, D \setminus \{\phi\})$ is empty, we need to show that $R(\phi, A \oplus B)$ is empty. We can show this by contradiction. Assume there is an $x^* \in R(\phi, A \oplus B)$. Since $(D \setminus \{\phi\}) \subseteq A \oplus B$, $x^* \in R(\phi, D \setminus \{\phi\})$. But we know that $R(\phi, D \setminus \{\phi\})$ is empty, so this cannot be.

To prove the second part, let $\omega = \text{argmax}_{\phi' \in A \oplus B} x \cdot \phi'$. The lemma is proved if we can show that $x \cdot \omega > x \cdot \omega'$ for all $\omega' \in W$. Let $(\alpha_1 + \beta_1) = \omega'$ for any $\omega' \in W$, $\alpha_1 \in A$ and $\beta_1 \in B$ and let $(\alpha + \beta) = \phi$ for $\alpha \in A$ and $\beta \in B$. By the conditions on $D$, we know that either $(\alpha_1 + \beta_1) \in D$, or $(\alpha_1 + \beta) \in D$, or $(\alpha + \beta_1) \in D$. Assume $(\alpha_1 + \beta) \in D$ (the other two cases are similar). Since $x \in R(\phi, D \setminus \{\phi\})$, $x \cdot \phi = x \cdot (\alpha + \beta) > x \cdot (\alpha_1 + \beta)$. This implies that $x \cdot \alpha > x \cdot \alpha_1$. Adding $\beta_1$ to both sides gives us that $x \cdot (\alpha + \beta_1) > x \cdot (\alpha_1 + \beta_1) = x \cdot \omega'$. By the definition of $\omega$, $x \cdot \omega \geq x \cdot (\alpha + \beta_1)$. Hence $x \cdot \omega > x \cdot \omega'$. The lemma follows.  □

Different choices of $D$ result in different incremental pruning algorithms. In general, the smaller the $D$ set, the more efficient the algorithm. Equation 13 is equivalent to using Monahan's (1982) filtering algorithm in INCPRUNE, Equation 15 is equivalent to using Lark's filtering algorithm (White 1991) in INCPRUNE (i.e., IP, as described earlier).

We refer to variations of the incremental pruning method using a combination of Equations 16 and 17 as the restricted region (RR) algorithm. Using either Equation 16 or 17 exclusively in the incremental pruning algorithm will improve the total constraint complexity of the algorithm to $O(|S^a|\sum_z |S_z^a|^2 + |S^a|^2|\mathcal{Z}|)$ constraints. Although the asymptotic total number of linear programs does not change, RR actually requires slightly more linear programs than IP in the worst case. However, empirically it appears that the savings in the total constraints usually saves more time than the extra linear programs require.

An even better variation of incremental pruning selects whichever $D$ set is smallest from among Equations 15, 16 and 17. This will usually yield a faster algorithm in practice, though it makes this variation much harder to analyze. The only extra work that is required is some bookkeeping to track how vectors were created and the sizes of the various sets that we will choose from.

In principle, it is also possible to choose a $D$ set that is the smallest set satisfying conditions 1 and 2. This appears to be closely related to the NP-hard vertex-cover problem; we are investigating efficient alternatives.

## 6  EMPIRICAL RESULTS

Although asymptotic analyses provide useful information concerning the complexity of the various algorithms, they provide no intuition about how well algorithms perform in practice on typical problems. Another shortcoming of these analyses is that they can hide important constant factors and operations required outside of the linear programs. To address these shortcomings, we have implemented IP and variations and have run them on a suite of test problems to gauge their effectiveness. All times given are in CPU seconds on a SPARC-10.

We profiled the execution and found that more than 95% of the total execution time was spent solving linear programs[4], verifying that the linear programs are the single most important contributor to the complexity of the algorithms.

To ensure fairness in comparison, we embedded all of the algorithms in the same value-iteration code and used as many common subroutines as possible. We also used a commercial linear programming package to maximize the stability and efficiency of the implementation.

We ran IP, RR, exhaustive and linear support algorithms on 9 different test problems listed in Table 1 (complete problem definitions are available at http://www.cs.brown.edu/people/arc/research/pomdp-examples.html). The "Stages" column reports the number of iterations of value iteration we ran and the "$|V_f|$" column indicates the number of vectors in the final value function[5].

Table 2 lists the total running time for each algorithm on each of the 9 test problems. The results indicate that RR works extremely well on a variety of test problems. We do not list run times for the linear support algorithm because, in all cases, it was unable to run to completion. This is because of memory limitations; space requirements for the linear support algorithm increase dramatically as a function of the number of states. We terminated algorithms that failed to complete in 8 hours (28800 seconds); as a result, the exhaustive algorithm ("Exh.") was only able to complete three of the test problems (all of which had only two observations). On the three small test problems the exhaustive algorithm was able to complete, it actually out performed all the other algorithms.

For all but two of the test problems, the witness algorithm was within a factor of 5 of the performance of RR. To highlight the advantage of the incremental-pruning-based algorithms, we chose the two test problems for which RR was more than 5 times faster than witness (Network, 8.9, and Shuttle, 11.5), and ran for a larger number of stages. As shown in Table 3, the witness algorithm is unable to solve a problem in 8 hours that RR can solve in 43 minutes (2621 seconds).

Although linear programming consumes most of the running time in the algorithms we examined, there are actually three phases of the value-iteration algorithm that contribute linear programs: finding the minimum-size $S_z^a$ sets, constructing the $S^a$ sets from the $S_z^a$ sets, and constructing $S'$ by combining the $S^a$ sets. Of these, only constructing the $S^a$ sets is different between witness, IP, and RR, so we have chosen to present the execution times in two ways. The first, as illustrated in Table 2 as $T_{\text{TOTAL}}$, represents the complete running time for all stages and all phases. The second, shown in Table 4 as $T_{S^a-\text{BUILD}}$, is the execution time over all stages that was devoted to constructing the $S^a$ sets from the $S_z^a$ sets.

As the data in Tables 2 and 4 show, IP performs better than the witness algorithm on all the test problems. These tables also show how difficult it is to analyze the exact amount of savings IP yields; the amount of savings achieved varies considerably across problems.

Table 2: Total execution time (sec.)

| Test Problem | $T_{\text{TOTAL}}$ | | | |
|---|---|---|---|---|
| | Witness | IP | RR | Exh. |
| 1D maze | 9.3 | 2.3 | 2.3 | 2.2 |
| 4x3 | 727.1 | 346.0 | 157.0 | >28800 |
| 4x4 | 3226.0 | 1557.0 | 909.2 | 216.7 |
| Cheese | 351.8 | 215.7 | 203.3 | >28800 |
| Part painting | 5608.4 | 4249.2 | 5226.4 | 1116.9 |
| Network | 6422.9 | 1066.6 | 722.5 | >28800 |
| Aircraft ID | 417.0 | 234.1 | 166.0 | >28800 |
| Shuttle | 1676.7 | 200.8 | 145.9 | >28800 |
| 4x3 CO | 24.6 | 22.8 | 22.7 | >28800 |

Table 3: Total execution time (sec.) for extended tests.

| Test Problem | Stages | $T_{\text{TOTAL}}$ | | |
|---|---|---|---|---|
| | | Witness | IP | RR |
| Network | 20 | >28800 | 4976.8 | 2621.3 |
| Shuttle | 9 | >28800 | 5121.3 | 2767.7 |

---

[4]This profiling data was computed running witness, IP, and RR on the 4x3 problem for 8 stages.

[5]The number of stages was determined by finding the maximum number of stages that the witness algorithm was able to complete within 7200 seconds. In some of the test problems, the witness algorithm found the optimal infinite-horizon value function in under 7200 seconds, so we picked the number of iterations required to converge to within machine precision of the optimal value function.

Table 1: Test problem parameter sizes.

| Test Problem | States | Acts. | Obs. | Stages | $|V_f|$ | Reference |
|---|---|---|---|---|---|---|
| 1D maze | 4 | 2 | 2 | 70 | 4 | |
| 4x3 | 11 | 4 | 6 | 8 | 436 | Parr & Russell (1995) |
| 4x3 CO | 11 | 4 | 11 | 367 | 4 | Russell & Norvig (1994) |
| 4x4 | 16 | 4 | 2 | 374 | 20 | Cassandra, Kaelbling, & Littman (1994) |
| Cheese | 11 | 4 | 7 | 373 | 14 | McCallum (1993) |
| Part painting | 4 | 4 | 2 | 371 | 9 | Kushmerick, Hanks, & Weld (1995) |
| Network | 7 | 4 | 2 | 14 | 438 | |
| Shuttle | 8 | 3 | 5 | 7 | 481 | Chrisman (1992) |
| Aircraft ID | 12 | 6 | 5 | 4 | 258 | |

Table 4: Total time (sec.) spent constructing $S^a$ sets.

| | $T_{S^a-\text{BUILD}}$ | | |
|---|---|---|---|
| Test Problem | Witness | IP | RR |
| 1D maze | 7.1 | <0.1 | <0.1 |
| 4x3 | 599.1 | 220.5 | 31.0 |
| 4x4 | 2252.6 | 644.4 | 0.9 |
| Cheese | 221.9 | 84.4 | 72.2 |
| Part painting | 5226.5 | 3834.4 | 4819.5 |
| Network | 5954.7 | 615.4 | 255.4 |
| Aircraft ID | 359.1 | 176.4 | 108.3 |
| Shuttle | 1566.2 | 92.5 | 38.1 |
| 4x3 CO | 2.6 | 0.9 | 0.9 |

For RR, the set $D$ was defined by Equation 16 if $|B| < |A|$ and Equation 17 otherwise; in most cases, this is equivalent to using the equation that leads to the smaller size for $D$. Looking at the data for RR, we see that in all but one case it is faster than IP. Again, the precise amount of savings varies and is difficult to quantify in general.

## 7 DISCUSSION & CONCLUSIONS

In this paper, we examined the incremental pruning method for performing dynamic-programming updates in partially observable Markov decision processes. Incremental pruning compares favorably in terms of ease of implementation to the simplest of the previous algorithms (exhaustive), has asymptotic performance as good as or better than the most efficient of the previous algorithms (witness), and is empirically the fastest algorithm of its kind for solving a variety of standard POMDP problems.

A complete incremental pruning algorithm (RR) is shown in Figure 4.

There are several important outstanding issues that should be explored. The first is numerical precision—

```
DP-UPDATE(S)
1  for each a in A
2    do for each z in Z
3      do S_z^a ← FILTER({τ(α, a, z)|α ∈ S_{t-1}})
4    S^a ← INCPRUNE(S_{z_1}^a, ..., S_{z_k}^a)
5  S' ← FILTER(∪_a S^a)
6  return S'

INCPRUNE(S_{z_1}^a, ..., S_{z_k}^a)
1  W ← RR(S_{z_1}^a, S_{z_2}^a)
2  for i ← 3 to k
3    do W ← RR(W, S_{z_i}^a)
4  return W

RR(A, B)
1  F ← A ⊕ B
2  W ← ∅
3  for each s in S
4    do ω ← argmax_{φ∈F} e_s · φ
5      W ← W ∪ {ω}
6      F ← F \ {ω}
7  while F ≠ ∅
8    do (α + β) ∈ F
9      D_1 ← ({α} ⊕ B) ∪ {α_1 + β|(α_1 + β) ∈ W}
10     D_2 ← (A ⊕ {β}) ∪ {β_1 + α|(α + β_1) ∈ W}
11     if |B| < |A|
12       then D ← D_1
13       else D ← D_2
14     x ← DOMINATE(α + β, D)
15     if x = ⊥
16       then F ← F \ {α + β}
17       else ω ← argmax_{φ∈F} x · φ
18         W ← W ∪ {ω}
19         F ← F \ {ω}
20  return W
```

Figure 4: Complete RR algorithm.

each of the algorithms we studied, witness, IP, and RR, have a precision parameter $\epsilon$, but the effect of varying $\epsilon$ on the accuracy of the answer differs from algorithm to algorithm. Future work will seek to develop an algorithm with a tunable precision parameter so that sensible approximations can be generated.

From a theoretical standpoint, there is still some work to be done developing better best-case and worst-case analyses for RR. This type of analysis might shed some light on whether there is yet some other variation that would be a consistent improvement over IP.

In any event, even the slowest variation of the incremental pruning method that we studied is a consistent improvement over earlier algorithms. We feel that this algorithm will make it possible to greatly expand the set of POMDP problems that can be solved efficiently.